\documentclass[11pt,a4paper]{article}
\usepackage[hyperref]{emnlp2020}
\usepackage{times}
\usepackage{latexsym}
\usepackage{amsmath}
\usepackage{multirow}
\usepackage{amssymb}
\usepackage{graphicx}

\usepackage{listings}
\usepackage{xspace}
\usepackage{hyperref}
\usepackage{makecell}
\usepackage{microtype}

\aclfinalcopy 
\def\aclpaperid{1054} 

\newcommand{\red}[1]{\textcolor{red}{#1}}
\newcommand{\blue}[1]{\textcolor{blue}{#1}}
\newcommand{\system}{RE-Flex\xspace}
\newcommand{\newparagraph}[1]{\paragraph{\textit{#1}}}

\DeclareMathOperator*{\argmax}{arg\,max}

\title{Unsupervised Relation Extraction from Language Models using Constrained Cloze Completion}

\author{Ankur Goswami, Akshata Bhat, Hadar Ohana, Theodoros Rekatsinas \\
  University of Wisconsin-Madison \\
  \texttt{\{agoswami6,abhat6,hohana,rekatsinas\}@wisc.edu} \\
  }

\date{}

\begin{document}
\maketitle
\begin{abstract}
We show that state-of-the-art self-supervised language models can be readily used to extract relations from a corpus without the need to train a fine-tuned extractive head. We introduce \system, a simple framework that performs constrained cloze completion over pretrained language models to perform unsupervised relation extraction. \system uses contextual matching to ensure that language model predictions matches supporting evidence from the input corpus that is relevant to a target relation. We perform an extensive experimental study over multiple relation extraction benchmarks and demonstrate that \system outperforms competing unsupervised relation extraction methods based on pretrained language models by up to 27.8 $F_1$ points compared to the next-best method. Our results show that constrained inference queries against a language model can enable accurate unsupervised relation extraction.
\end{abstract}

\section{Introduction}\label{sec:intro}

Relation extraction is a fundamental problem in constructing knowledge bases from unstructured text. The goal of relational extraction is to identify mentions of relational facts (i.e., binary relations between entities) in a text corpus. Traditionally, relation extraction systems leverage supervised machine learning approaches to train specialized extraction models for different relations~\cite{dong2014knowledge, shin2015incremental}. However, advances in natural language understanding models, such as BERT~\cite{devlin_bert:_2018} and RoBERTa \cite{liu2019roberta}, have shifted the focus towards {\em general relation extraction} where a single natural language model is used for extraction across different relations~\cite{levy_zero-shot_2017}.

A key idea behind general relation extraction is to leverage question answering (QA) models and use the reading comprehension capabilities of modern natural language models to identify relation mentions in text. For example, the relation {\texttt{drafted\_by}} can be completed  for the subject {\texttt Stephen Curry} by answering the question {\texttt{Who drafted Stephen Curry?}} State-of-the-art results leverage fine-tuned QA models over self-supervised contextual representations~\cite{devlin_bert:_2018, radford2018improving}. Initial approaches~\cite{levy_zero-shot_2017} learn {\em extractive QA models} by exploiting annotated question-answer pairs and following a supervised setting. 

While effective in domains related to the annotated question-answer data, supervised extractive QA approaches can fail to generalize to new domains for which annotations are not available \cite{dhingra_simple_2018}. For this reason, more recent approaches~\cite{lewis_unsupervised_2019} propose to use  automatically generated question-answer pairs for training and adopt a weakly-supervised setting~\cite{lewis_unsupervised_2019}.  However, noisy or inaccurate training data leads to a significant drop in performance.

In this work, we revisit the problem of general relation extraction and show that one can perform unsupervised relation extraction by directly using the generative ability of self-supervised contextual language models and without training a fine-tuned QA model. We build upon the recent observation that modern language models encode the semantic information captured in text and are capable of generating answers to relational queries by answering cloze queries that represent a relation~\cite{petroni-etal-2019-language}. For instance, the previous extraction example can be transformed to the cloze query {\texttt{Stephen Curry was drafted by [MASK]}} and the language model can be used to predict the most probable value for the masked token. Further, recent works~\cite{radford2019language, petroni2020how} show that prefixing cloze queries with relevant information, i.e., \emph{relevant context}, can improve extraction accuracy by utilizing the models' reading comprehension ability \cite{radford2019language, petroni2020how}. While promising, we show that an out-of-box application of these methods to general relation extraction falls short of extractive QA models. The core limitation is that of {\em factual generation}: language models do not memorize general factual information \cite{petroni-etal-2019-language}, and are liable to predict off-topic or non-factual tokens~\cite{see_get_2017}.

We propose a novel two-pronged approach that ensures factual predictions from a contextual language model. First, given an extractive relational cloze query and an associated context, we propose a method to restrict the model's answer to the query to be factual information in the associated context. We introduce a context-constrained inference procedure over language models and does not require altering the pre-training algorithm. This procedure relies on redistributing the probability mass of the language model's initial prediction to tokens only present in the context. By restricting the model's inference to be present in the context, we ensure a factual response to a relational cloze query. This strategy is similar to methods used in unsupervised neural summarization \cite{zhou_simple_2019} to ensure factual summary generation. Second, we introduce an unsupervised solution to determining whether the context associated with the query contains an answer to a relational query. We propose an information theoretic scoring function to measure how well a relation is represented in a given context, then cluster contexts into ``accept" and ``reject" categories, denoting whether the contexts express the relation or not. 

We present an extensive experimental evaluation of \system against state of the art general relation extraction methods across several settings. We demonstrate that \system outperforms methods that rely on weakly supervised QA models~\cite{dhingra_simple_2018, lewis_unsupervised_2019} by up to $27.8$ $F_1$ points compared to the next-best method, and can even outperform methods that rely on supervised QA models~\cite{levy_zero-shot_2017} by up to $12.4$ $F_1$ points in certain settings. Our results demonstrate that by constraining language generation, \system yields accurate unsupervised relation extractions.
\begin{figure*}[t]
\center
\includegraphics[width=\textwidth]{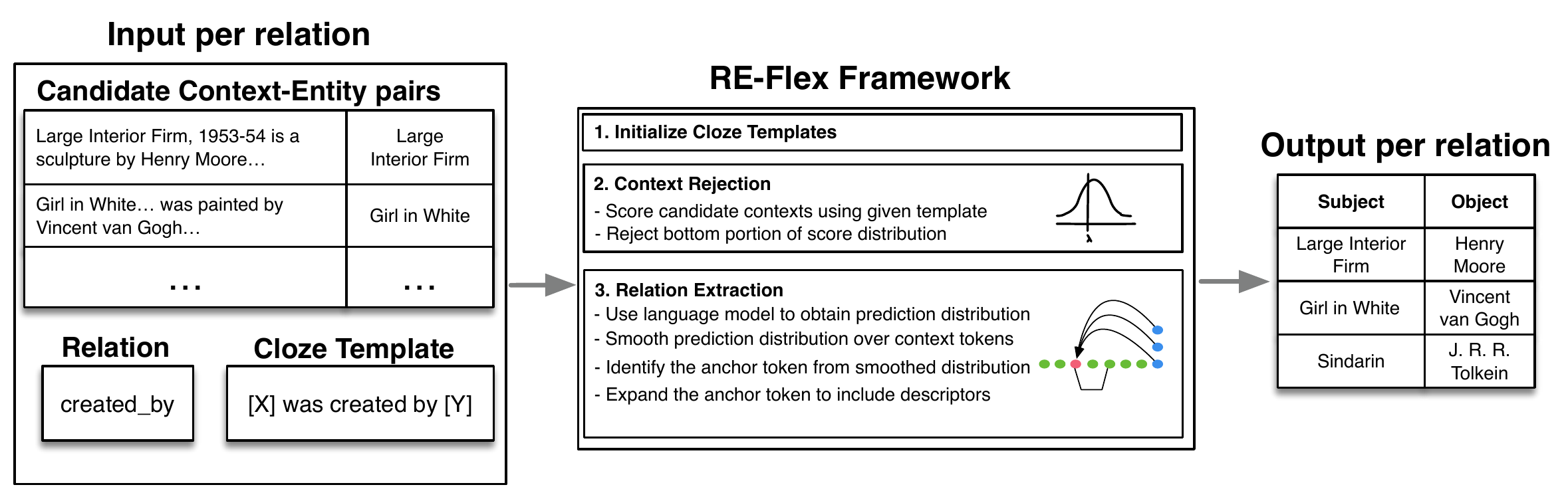}
\caption{The \system Framework Overview}
\centering
\label{fig:overview}
\end{figure*}

\section{Related Work}
Typical relation extraction relies on rule-based methods~\cite{Soderland:1995:CIC:1643031.1643069} and supervised machine learning models that target specific relation types~\cite{Hoffmann:2011:KWS:2002472.2002541, dong2014knowledge, shin2015incremental}. These approaches are limited to predefined relations and do not extend to relations that are not specified during training. To alleviate this problem, open information extraction (OpenIE)~\cite{banko2007open} proposes to represent relations as unstructured text. However, in OpenIE different phrasings of the same relation can be treated as different relations, leading to redundant extractions. To address this issue, Universal Schema~\cite{riedel2013relation} uses matrix factorization to link OpenIE relations to an existing knowledge base to distill extracted relations. Our problem is aligned with the thrust of OpenIE: enabling general relations to be extracted from text corpora without relation specific supervision.

More recently, question answering has become a popular method to extract relations from text. \citet{levy_zero-shot_2017} showed that casting relation extraction as a QA problem can enable new, unseen relations to be extracted without additional training. Advances in large self-supervised language models \cite{radford2018improving} have enabled QA models to achieve human level performance on some datasets~\cite{rajpurkar-etal-2016-squad}. Because these models are trained on a slot-filling objective, there has been a branch between methods that use a QA head to extract spans from input, and methods that use token generation capability of language models to perform information extraction. Both are relevant to our work.

Many QA-based methods have been proposed to identify spans from text. \citet{das_building_2018} present a reading comprehension model based on the architecture of \citet{chen_reading_2017} to track the dynamic state of a knowledge graph as the model reads the text. \citet{li_entity-relation_2019} proposes a multi-turn QA system to extract relational fact triplets. \citet{xiong-etal-2019-improving} map evidence from a knowledge base to natural language questions to improve performance in the general QA setting. Most relevant QA systems to our work are the works of \citet{lewis_unsupervised_2019} and \citet{dhingra_simple_2018}, which propose weak supervision algorithms to generate QA pairs over new corpora for training. We compare to these models in our experiments.

There are also many generative methods that rely only on a language model to generate the answer to queries. \citet{radford2019language} show that self-supervised language models can generate answers to questions. \citet{petroni-etal-2019-language} show that given natural language cloze templates that represent relations, masked language models \cite{devlin_bert:_2018} can answer relational queries directly. \citet{petroni2020how} extends on this work to show that retrieving factual evidence to associate with relation queries can further benefit answer generation. \citet{logan-etal-2019-baracks} present a knowledge graph language model that can choose between outputting tokens from a base vocabulary, or entities from a linked knowledge base. \citet{bosselut2019comet} show that language models can generate commonsense knowledge bases if pretrained on another corpus and fine-tuned on a commonsense knowledge base. We build on this work, but choose to focus on formulating an improved inference procedure for generative query answering, instead of focusing on learning better representations or using out of the box inference.

\section{Problem Statement}\label{sec:problem}
We consider a slot filling form of relation extraction: given incomplete relations, we must complete the relations using evidence from an underlying text source. We assume a set of input relations $R$. For each relation $r \in R$, we assume access to a collection of entity-context candidate pairs. Let $EC_r$ denote this collection for relation $r$. We consider each pair $(e, c) \in EC_r$ to be candidate evidence that some span in $c$ completes relation $r$ for the given entity mention $e$. If we consider the context to be composed of a sequence of tokens $c = (c_1, c_2, \dots, c_n)$, we must return some subsequence $a = (c_i, ..., c_{i+m})$ such that the relation $r(e, a)$ holds, or $\varnothing$ if $c$ does not express the relation for the given entity.

Furthermore, we represent each relation with a {\em cloze template}: a natural language representation of what the relation is attempting to capture. A cloze template for relation $r$ is a sequence of tokens $t = (t_1, \dots, t_{sub}, \dots, t_{obj}, \dots, t_{k})$, where $t_{sub}$ and $t_{obj}$ are special tokens denoting the expected locations of the subject and object entities of the relation. For each $(e, c)$, we substitute the special token $t_{sub}$ with $e$. Let $t(e) = (t_1, \dots, e, \dots, t_{obj}, \dots, t_{k})$ denote this substitution. We form our final cloze query by concatenating the context $c$ to the cloze task $t(e)$ and denote the close query $q(e,c) = [c, t(e)]$. 

Given a cloze query $q(e,c)$, we express relation extraction as the following inference task: predict if there is a subsequence of the context $c$ that correctly substitutes the special token $t_{obj}$ in the cloze task $t(e)$, otherwise return $\varnothing$. As an example, consider the relation \texttt{drafted\_by}. An example candidate entity-context pair in the pair set for of relation is (\texttt{Stephen Curry}, \texttt{The Warriors drafted Steph Curry.}). Using the relational template \texttt{$t_{sub}$ was drafted by $t_{obj}$}, we form our full cloze query for the pair: \texttt{The Warriors drafted Steph Curry. Stephen Curry was drafted by $t_{obj}$}.

\section{The RE-Flex Framework}

An overview of \system is shown in Figure~\ref{fig:overview}. Given a target relation, \system assumes as input a set of entities, a set of candidate contexts, and a cloze template expressing the relation. The output of \system is a table containing subject-object instances of this relation for the input entities. \system is built around two key parts: 1) context rejection and 2) anchor token identification and token expansion. In the first part, \system determines if the cloze query for a candidate entity-context pair does not contain a valid mention of the target relation, and hence, we must return $\varnothing$. In the second part, given valid entity-context pairs for the target relation, \system identifies the subsequence in the corresponding context that completes the relation for the given entity. We describe each part next.




\subsection{Context Rejection}\label{sec:context_rejection}
For each relation $r$, we must determine which of the candidate pairs $(e, c) \in EC_r$ express relation $r$ for entity $e$, and return $\varnothing$ for those that do not. The problem can be naturally considered as a clustering problem, where we group  elements of $EC_r$ into an accept cluster $I_{c}$ or a reject cluster $I_{-c}$. Given this regime, we must develop a general method to determine how well a given entity-context pair $(e, c)$ expresses a target relation. Using the natural language representation of the relation, we formulate a scoring function to measure how much each context expresses the relation. We then determine a threshold on these scores to partition the pairs. 

We propose the following mechanism: First, we leverage the fact that the cloze template $t$ for a target relation $r$ is the natural language representation of the relation and assume that it captures the intention of the relation. We formulate a scoring function $f(c, t(e))$ which takes as input a context $c$ and $t(e)$---the cloze template where we have substituted $t_{sub} = e$---and returns a measurement of how well each token in the template is captured in a given context. Second, for some threshold $\epsilon$, if $f(c, t(e)) > \epsilon$, we assign the corresponding pair $(e, c)$ to $I_{c}$, and to $I_{-c}$ otherwise.

We design $f$ with the following intuition: if each word in the template co-occurs many times with any word in the context, the relation is likely to be expressed. We define $f$ as follows:

$$
f(c, t(e)) = \frac{1}{|t(e)|}\sum_{i=0}^{|t(e)|} \max_{j \in [1, |c|]} \text{PMI}(t(e)[i], c[j])
$$

where PMI is the Pointwise Mutual Information \cite{church1990word}, $|t(e)|$ and $|c|$ are the total number of tokens in the cloze task $t(e)$ and the context $c$ respectively, $t(e)[i]$ denotes the token in position $i$ of the cloze task $t(e)$, and $c[j]$ denotes the token in position $j$ of context $c$. For two words $x$ and $y$, PMI is defined as $\text{PMI}(x, y) = \log \frac{p_q(x, y)}{p(x)p(y)}$, where $p_q(x, y)$ is the probability that $x$ and $y$ co-occur in a $q$-gram in the corpus and $p(x)$ is the marginal probability of $x$ occurring in the corpus; we set $q =5$. 

We estimate PMI using the cosine similarity between the word embeddings produced by optimizing the skip-gram objective over a target corpus \cite{mikolov2013efficient}. This approach does not suffer from missing values in the PMI matrix, as an empirical estimate of the PMI matrix might \cite{levy2014neural}.  As proven in \citet{arora_latent_2016}, for two words $x$ and $y$ and their word embeddings $v_x \in \mathbb{R}^d$ and $v_y \in \mathbb{R}^d$ we have that:

$$\text{PMI}(x, y) \approx \frac{\langle v_x, v_y \rangle}{||v_x||||v_y||}$$

We use a simple inlier detection method to determine the threshold $\epsilon$. We assume that entity-candidate contexts for each relation $r$ are relatively well-aligned, i.e., the majority of elements in $EC_r$ contain a true mention of relation $r$ for the entity associated with each element. Let $Q_r$ denote the set of all possible correct entity-context pairs for $r$. We assume that for any valid pair $(e,c)$ the score $f(c, t(e))$ follows a normal distribution $\mathcal{N}(\mu_r, \sigma^2_r)$, and hence, we expect that for most entity-context pairs the similarity scores to the cloze task associated with the relation will be centered around the mean $\mu_r$. Given the above modeling assumptions, we estimate $\mu_r$ and $\sigma^2_r$ as follows:

\begin{align*}
&\mu_r = \frac{1}{|EC_r|}\sum_{(e,c) \in EC_r} f(c, t(e)) \\
&\sigma_r^2 = \frac{1}{|EC_r|}\sum_{(e,c) \in EC_r} (f(c, t(e)) - \mu_t)^2
\end{align*}

We then let $\epsilon$ is $\epsilon = \mu_r - \lambda \sigma_r$ where $\lambda$ is a hyperparameter. We assign all $(e_r, c_r)$ pairs to $I_c$ if $f(c_r, t_r) > \epsilon$, and assign the rest to $I_{-c}$. For all pairs in $I_{-c}$, we return $\varnothing$.
\subsection{Relation Extraction} \label{sec:alignment}
We discuss how \system performs relation extraction given a valid entity-pair context. For this part, we assume access to a pre-trained contextual language model---in \system we use RoBERTa \cite{liu2019roberta}. For a valid entity-context pair $(e, c)$ for relation $r$, we construct the cloze query $q(e,c) = [c, t(e)]$ by replacing the subject mask token $t_{sub}$ in the cloze template $t$ with $e$, and given the sequence $q(e,c)$ we identify the token span $\alpha$ in $c$ that should replace the object  mask token $t_{obj}$ in $t(e)$ to complete relation $r$ for entity $e$. 

At a high-level, we follow the next process to identify span $\alpha$: first, we consider the raw predictions of the pre-trained model for $t_{obj}$, and smooth the scores of these predictions by restricting valid predictions to correspond only to tokens present in the context $c$; we pick the context token with the highest final score, which we refer to as the \emph{anchor token}. Second, given the anchor token in $c$, we return an expanded span from $c$ that contains descriptors of the anchor token. We describe each of these two steps next. 

\paragraph{Anchor token identification} We focus on the first step described above. Given an entity-context pair $(e,c)$ that contains a true mention of relation $r$, the desired answer to the cloze query $q(e,c)$ corresponds to a span of tokens $\alpha$ in $c$. The task of anchor token identification is to identify any token in span $\alpha$. To identify such a token, we constraint the inferences of the pre-trained model to tokens in the context $c$.

Given the cloze query $q(e,c) = [c, t(e)]$, also denoted hereafter $q$ for simplicity, we first use the pre-trained model, denoted hereafter by $M$, to obtain a prediction for the masked token $t_{obj}$ (see Section~\ref{sec:problem}). Let $V$ denote the vocabulary of all tokens present in the domain of consideration. For each token $v \in V$, we can use $M$ to obtain a probability that $v$ should be used to complete the masked token $t_{obj}$. Let $p_{q,M}(v) = p(t_{obj}=v;q, M)$ denote this probability for token $v$. 

To obtain a factual prediction, we reassign the above probability mass to only to the tokens found in context $c$. We leverage the contextual model $M$ for this step. For the token at each position in the context sequence $c$, we find all tokens in $V$ that are semantically compatible with it, given the cloze query $q(e,c)$, and reassign the probability mass of these tokens proportionally. Consider the $i$-th position in the context $c$. We define the new probability mass for token $c[i]$, denoted by $z_{q,M}(c[i])$, as:

$$
z_{q,M}(c[i]) = \sum_{v \in V}p_{q,M}(v)\cdot D(c[i],v)
$$

where $D(c[i],v)$ is a non-negative normalized score indicating the semantic compatibility between tokens $c[i]$ and $v$. We have:

$$
D(c[i],v) = \frac{\exp(d(c[i],v))}{\sum_{j=1}^{|c|}\exp(d(c[j], v))}
$$

where the unnormalized scores $d(c[i],v)$ are obtained using the similarity between contextual embeddings obtained by model $M$. 

We define this contextual similarity more formally. Let $q^{e,c}(v)$ be the sequence corresponding to the cloze query $q(e,c)$ after we replace the masked object token $t_{obj}$ in the cloze template of the target relation with some token $v \in V$. That is for context $c = \{c_1, c_2, \dots, c_n\}$, entity $e$, and the cloze template $t = \{t_1, \dots, t_{sub}, \dots, t_{obj}, \dots, t_{m}\}$, we have $q^{e,c}(v)= \{c_1, \dots, c_n, t_1, \dots, e, \dots, v, \dots, t_{m}\}$. Given model $M$ and sequence $q^{e,c}(v)$, let $M(q^{e,c}(v))[k] \in \mathbb{R}^d$ be the contextual embedding returned by $M$ for the token at the $k$-th position of sequence $q^{e,c}(v)$. We define the unnormalized score $d(c[i], v)$ as:

$$
d(c[i],v) = \cos(M(q^{e,c}(v))[i], M(q^{e,c}(v))[obj])
$$

where $\cos(A,B)$ denotes the cosine similarity between two vectors, and $obj$ denotes the position of object token set to $v$ in sequence $q^{e,c}(v)$.

An exact computation of $z_{q,M}(c[i])$ would require $|V|$ forward passes. Instead, we propose to approximate $z_{q,M}(c[i])$. In practice, the language model's output distribution over the vocabulary has low entropy. Thus, we expect $p_{q,M}(v)$ to be zero for most $v \in V$. Therefore, we can approximate $z_{q,M}(c[i])$ by only summing over the top-$k$ tokens for the probability mass $p_{q,M}$. We define a set of proposal tokens $\tilde{V}$ to be these top-$k$ tokens. Empirically, we find that filtering out punctuation from $\tilde{V}$ also increases performance. We take the position of the anchor token in $c$, denoted by $a_{out}$ to be: 

$$
a_{out} = \argmax_{i \in \{1, \dots, |c|\}} \sum_{v \in \tilde{V}} p_{q,M}(v)\cdot D(c[i],v)
$$

This approximation only requires $k+1$ forward passes (one additional forward pass is needed to obtain the initial $p_{q,M}$ distribution) to compute the final prediction. We examine the effect of setting different $k$ in Appendix \ref{sec:app:micro}.

\paragraph{Anchor token expansion}
We use a simple mechanism to expand the single-token anchor to a multi-token span. Given an off-the-shelf named entity recognition (NER) model, we do the following: if the anchor word is within a named entity, return the entire entity. Otherwise, return just the anchor word. While this approach allows us to support multi-token answers, its quality is highly correlated to that of the NER model. In practice, we do not find this to be a limiting factor because most entities tend to span few tokens. We experimentally evaluate the effect of using NER to obtain multi-token spans in Appendix \ref{sec:app:micro}. We choose this approach as our focus is on studying if language models can be used directly for relation extraction.

\section{Experimental Evaluation}\label{sec:experiments}
We compare \system against several competing relation extraction methods on four relation extraction benchmarks. The main points we seek to validate are: (1) how accurately can \system extract relations by utilizing contextual evidence, (2) how does \system compare to different categories of extractive models.

\subsection{Experimental Setup}\label{sec:exp_setup}
We describe the benchmarks, metrics, and methods we use in our evaluation. We discuss implementation details in Appendix \ref{sec:app:competingdetails}.

\subsubsection{Datasets and Benchmarks}
We consider four relation extraction benchmarks. The first two, T-REx \cite{elsahar_t-rex:_nodate} and Google-RE\footnote{\url{ https://code.google.com/archive/p/ relation- extraction- corpus/}}, are datasets previously used to evaluate unsupervised QA methods \cite{petroni2020how}, and are part of the LAMA probe \cite{petroni-etal-2019-language}. We also consider the Zero-Shot Relation Extraction (ZSRE) benchmark \cite{levy_zero-shot_2017}, which is a dataset originally used to show that reading comprehension models can be extended to extractions of unseen relations. Finally, we adapt the TAC Relation Extraction Dataset (TACRED) \cite{zhang2017tacred} to the slot filling setting utilizing a protocol similar to that used in \citet{levy_zero-shot_2017}. We present the adaptation procedure, as well as a full table of benchmark characteristics in Appendix \ref{sec:app:datasets}. For the T-REx and Google-RE datasets \emph{all} inputs correspond to entity-\-context pairs that contain a valid relation mention. On the other hand, ZSRE and TACRED contain invalid inputs for which the extraction models should return $\varnothing$. We refer to the first two datasets as the LAMA benchmarks, while the latter two are general relation extraction benchmarks.

\subsubsection{Metrics} We follow standard metrics from Squad 1.0~\cite{rajpurkar-etal-2016-squad} and evaluate the quality of each extraction using two metrics: {\em Exact Match (EM)} and $F_1$-score. Exact match assigns a score of $1.0$ when the extracted span matches exactly the ground truth span, or $0.0$ otherwise. $F_1$ treats the extracted span as a set and calculates the token level precision and recall. For each relation, we compute the average EM and $F_1$ scores and then average these scores across relations.

\subsubsection{Defining cloze templates} We manually define cloze templates for each relation. As in previous work that explore language generation to complete knowledge queries \cite{petroni-etal-2019-language}, we note that these templates may not produce the optimal extractions. Moreover, we point out that subtle variations in cloze templates can cause variation in performance. As we report results that are averaged across many relations, error due to cloze definition is part of the end-to-end performance for the relevant methods.


\subsubsection{Competing Methods}
We consider three classes of competing methods: 1) models that rely on the generative ability of language models, 2) weakly-supervised QA models trained on an aligned set of question-answer pairs, and 3) supervised QA models trained on annotated question-answer pairs. Implementation details are found in Appendix \ref{sec:app:competingdetails}.

\paragraph{Generative Methods} We compare to the naive cloze completion (NC) method of \citet{petroni-etal-2019-language}, which queries a masked language model to complete a cloze template representing a relation, without an associated context. We also consider the method of \citet{petroni2020how} (GD), which concatenates the context to the cloze template, and greedily decodes an answer to the relational query. This method is the same as that used in \citet{radford2019language} to show language models are unsupervised task learners. We use the RoBERTa language model \cite{liu2019roberta} for both these baselines.

\paragraph{Weakly-supervised QA Methods} We compare against two proposed weakly-supervised QA methods. The first method  \cite{lewis_unsupervised_2019} (UE-QA) uses a machine translation model to create questions from text using an off-the-shelf NER model, then trains a question answering head on the generated data to extract spans from text. The second method \cite{dhingra_simple_2018} (SE-QA) is a semi-supervised approach to QA. It also uses an NER model to generate cloze-style question-answer pairs and then trains a QA model on these pairs. Authors provide generated data for both methods, which we use to train a BERT-Large QA model \cite{devlin_bert:_2018}.

\paragraph{Supervised QA Methods} Finally, we compare against three supervised QA models trained on annotated question-answer pairs. We train BiDAF \cite{seo2016bidirectional}, extended to be able to predict no answer \cite{levy_zero-shot_2017} on Squad 2.0 \cite{rajpurkar-etal-2018-know}.  Additionally, we train BERT-Large on Squad 2.0 (B-Squad) and the training set of ZSRE (B-ZSRE). For Google-RE and T-REx, we do not allow these models to return $\varnothing$. These baselines require the existence of a significant number of human annotations in the case of Squad2.0, or the existence of a large reference knowledge base in the case of ZSRE.

\begin{table*}[t]
\small
\centering
\begin{tabular}{c|c |c|c|c c | c c | c c c}
\hline
\multicolumn{1}{c|}{\multirow{2}{*}{\small{\textbf{Category}}}} &\multicolumn{1}{c|}{\multirow{2}{*}{\small{\textbf{Dataset}}}}& \multicolumn{1}{c|}{\multirow{2}{*}{\small{\textbf{Metric}}}} & \multicolumn{1}{c|}{\multirow{2}{*}{\small{\textbf{\system}}}} & \multicolumn{2}{c|}{\small{\textbf{Generative}}}& \multicolumn{2}{c|}{\textbf{Weakly-supervised}} & \multicolumn{3}{c}{\small{\textbf{Supervised}}}\\ 
\cline{5-11}& & & & \small{\textbf{NC}} & \small{\textbf{GD}} & \small{\textbf{UE-QA}} & \small{\textbf{SE-QA}} & \small{\textbf{BiDAF}} & \small{\textbf{B-Squad}} & \small{\textbf{B-ZSRE}} \\ \hline
 \multirow{4}{*}{\small{LAMA}} & \multirow{2}{*}{\small{T-REx}} & \small{EM} & \textbf{56.0} & \red{22.9} & \red{43.2} & \red{14.2} & \red{21.2} & \red{35.5} & \red{42.2} & \red{46.2}\\
& & \small{$F_1$} & \textbf{56.0} & \red{22.9} & \red{43.7} & \red{19.3} & \red{28.2} & \red{46.6} & \red{52.0} & \red{52.7}\\
\cline{2-11} & \multirow{2}{*}{\small{Google-RE}} & \small{EM} & \textbf{87.2} & \red{5.6} & \red{75.6} & \red{51.9} & \red{19.7} & \red{53.9} & \red{52.9} & \red{70.5}\\
& & \small{$F_1$} & \textbf{87.2} & \red{5.6} & \red{75.7} & \red{65.0} & \red{25.3} & \red{74.8} & \red{76.5} & \red{75.8}\\ \hline
\multirow{4}{*}{\small{GR}} & \multirow{2}{*}{\small{ZSRE}} & \small{EM} & 43.7 & \red{14.1} & \red{27.7} & \red{16.7} & \red{17.3} & \red{40.2} & \blue{66.5} & \blue{\textbf{82.0}}\\
& & \small{$F_1$} & 46.9 & \red{14.9} & \red{30.7} & \red{23.7} & \red{24.9} & \blue{49.0} & \blue{74.1} & \blue{\textbf{84.8}}\\
\cline{2-11} & \multirow{2}{*}{\small{TACRED}} & \small{EM} & 49.4 & \red{9.0} & \red{25.6} & \red{17.5} & \red{13.9} & \red{49.3} & \blue{\textbf{56.9}} & \blue{54.4}\\
& & \small{$F_1$} & 50.1 & \red{9.0} & \red{26.3} & \red{22.0} & \red{18.6} & \blue{53.7} & \blue{\textbf{61.1}} & \blue{55.3}\\

\hline
\end{tabular}
\caption{Performance for all methods on the four benchmarks. Datasets are divided into two categories, LAMA and General Relation (GR) denoting whether the dataset requires that $\varnothing$ be returned for any examples. Bold values highlight the best method. \red{Red} and \blue{blue} values denote worse or better performance than RE-Flex, respectively.}
\label{tab:overall_results}
\end{table*}

\subsection{End-to-end Comparisons}\label{sec:overall_performance}
We evaluate the performance of \system against all competing methods for different benchmarks. The results are shown in Table~\ref{tab:overall_results}.

\subsubsection{LAMA Benchmarks}
We focus on the LAMA benchmarks, which consist of the Google-RE and T-REx datasets. For these benchmarks, the context always contains the answer to the relational query, and the answer is a single token. We analyze the performance of \system against each group of baselines.

\paragraph{Comparison to Generative Methods} We first compare the performance of \system to that of the generative methods NC and GD. We see that \system outperforms NC by $33.1$ $F_1$ on T-REx and $81.6$ $F_1$ on Google-RE. We see that GD also outperforms NC. This observation suggests that retrieving relevant contexts and associating them with relational queries significantly increases the performance of generative relation extraction methods, as opposed to relying on the model's memory.

Compared to GD, \system shows an improvement of $12.3$ $F_1$ on T-REx and $11.5$ $F_1$ on Google-RE. We attribute this gain on \system's ability to constrain the language model's generation to tokens only present in the context.

\noindent{\textbf{Takeaway:}} Restricting language model inference ensures more factual predictions, and is key to accurate relation extraction when using the contextual language model directly.



\paragraph{Comparison to Weakly-supervised Methods}

We compare \system to UE-QA and SE-QA, which both construct a weakly-aligned noisy training dataset and fine-tune an extractive QA head on the produced examples. \system outperforms both approaches, yielding improvements of $27.8$ $F_1$ on T-REx and $22.2$ $F_1$ on Google-RE compared to the best performing method for each dataset.

Additionally, we see that, on these benchmarks, GD (despite yielding worse results than \system) also outperforms UE-QA and SE-QA. This result suggests that training on noisy training data can severely hamper downstream performance.

\noindent{\textbf{Takeaway:}} Using weak-alignment to train a QA head often leads to poor results, and it is better to use the model's generative ability instead. Below, we show that this behavior extends to general relation extraction benchmarks.



\paragraph{Comparison to Supervised Methods}
We find the surprising result that \system is better than all supervised methods. We believe the results can be attributed to the fact that the language model is able to capture the subset of relations in these datasets quite well. This finding is also supported by the fact that GD also yields comparable accuracy to the supervised methods.

As we examine below, this behavior is not as pronounced when considering the general relation extraction setting. Still, we are able to assert that for specific relation subsets, our inference procedure is able to outperform standard QA models.

\noindent{\textbf{Takeaway:}} Our findings strongly support that contextual models capture certain semantic relations \cite{petroni-etal-2019-language, petroni2020how}, but to outperform the performance of supervised models we still need \system's fine-tuned inference procedure. 



\begin{table}[t]
\centering
\small
\begin{tabular}{c|c c|c c} 
\multirow{2}{*}{\textbf{Method}} & \multicolumn{2}{c|}{\textbf{ZSRE}} & \multicolumn{2}{c}{\textbf{TACRED}} \\ \cline{2-5}
& \textbf{EM} & \textbf{$\mathbf{F_1}$} & \textbf{EM} & \textbf{$\mathbf{F_1}$} \\ \hline
Without rejection & 40.0 & 43.4 & 39.1 & 39.5 \\ 
With rejection & 43.7 & 46.9 & 49.4 & 50.1 \\ 
\end{tabular}
\caption{Context rejection ablation on general relation benchmarks.}
\label{tab:context_ablation}
\end{table}

\subsubsection{General Relation Benchmarks}

We now focus on ZSRE and TACRED, which are more reflective of our problem statement. Here, we must assert whether a candidate context contains a true expression of the relation, and produce multiple token spans as answers.

\paragraph{Comparison to Generative and Weakly-supervised Methods} We see that \system significantly outperforms all generative and weakly-supervised methods on these benchmarks. We outperform the next best method by $22.0$ $F_1$ on ZSRE and by $28.1$ $F_1$ on TACRED. In this realistic context, using the contextual language model without fine-tuning the corresponding inferences falls short, while a noisily trained QA head also exhibits poor performance. To understand if these results are to be attributed to \system's ability to reject contexts, we ablate the performance of \system with and without enabling context rejection (Section  \ref{sec:context_rejection}). The results are shown in Figure \ref{tab:context_ablation}. We see that context rejection leads to increased performance. For example, in TACRED it boosts \system's $F_1$ score by more than 10 points. We also see that even without the context rejection, \system  outperforms these classes of methods by up to $13.2$ $F_1$ compared to the next best method. This finding suggests that the combination of fine-tuned inference and context rejection leads to good performance.

\noindent{\textbf{Takeaway:}} In addition to restricted inference, incorporating context rejection is necessary for the general relation extraction setting. This finding is consistent with that for the LAMA benchmarks.




\paragraph{Comparison to Supervised Methods} We compare to supervised QA baselines on the general relation extraction benchmarks. Here, all competing approaches are trained on human annotated QA pairs. We find that \system performs comparably to BiDAF but falls short of the fine-tuned BERT-based QA models. Recall that BiDAF relies on a simpler attention-flow model, and does not use self-supervised language representations, as BERT does. The best performing BERT baselines see an average improvement of $37.9$ $F_1$ on ZSRE and $10$ $F_1$ on TACRED compared to RE-Flex. However, as we show next, there is a significant number of relations for which \system outperforms the BERT-based baselines for even up to $40$ $F_1$ points in TACRED and up to $60$ $F_1$ points in ZSRE.

\begin{figure}
\center
\includegraphics[width=0.48\textwidth]{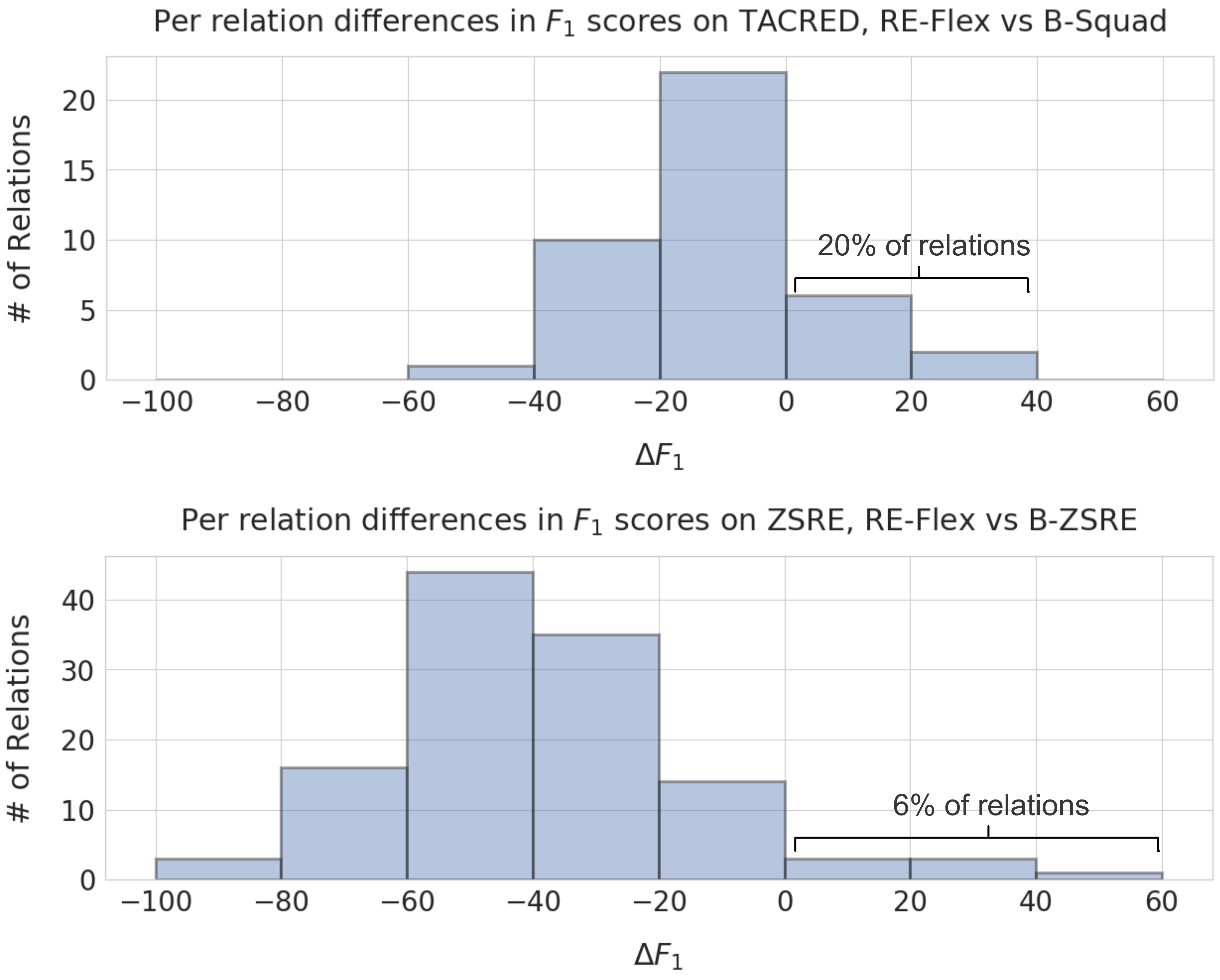}
\caption{Histogram breakdown of differences between $F_1$ performances between \system and the best performing supervised methods for each of the ZSRE and TACRED benchmarks. We see that for many cases the unsupervised approach of \system outperforms the fully-supervised BERT-based baselines.}
\centering
\label{fig:deltaf1}
\end{figure}

To better understand \system's behavior beyond the averaged $F_1$, we record the difference in $F_1$ scores between \system and each BERT baseline on a per relation basis. Histograms of these results can be found in Figure \ref{fig:deltaf1}. On TACRED, RE-Flex outperforms the best method for $20\%$ of relations and comes within $20.0$ $F_1$ for $26\%$ of relations. For ZSRE, RE-Flex outperforms the best method for $6\%$ of relations, and comes within $20.0$ $F_1$ for another $12\%$ of relations. These results show that for certain relations, RE-Flex can perform competitively or even better with supervised methods. 

We note that the relations for which \system performs better than the baselines tend to be simple many-to-one relations which are likely to be clearly stated in succinct ways. For example, \system outperforms baselines on the \texttt{cause\_of\_death} and \texttt{religious\_affiliation} relations. \system tends to fail on domain specific relations, such as \texttt{located\_on\_astronomical\_body}. Here, questions can incorporate specific output requirements (e.g., ``where" questions should return a location), and supervised models can learn these signals, whereas incorporating intention into language generation is an open research problem \cite{keskar2019ctrl}.

Finally, we note the performance drop of B-ZSRE when applied to the TACRED dataset. Both QA models perform similarly on TACRED, which does not have a QA training set associated with it. This shows that supervised QA models exhibit some bias towards the underlying corpus they are trained on, which supports claims in previous work \cite{dhingra_simple_2018}. We further expand on this result in Appendix \ref{sec:app:qabias}.

\noindent{\textbf{Takeaway:}} We find evidence that, for simple many-to-one relations, fine-tuned inference over self-supervised models can exhibit comparable or better performance than fine-tuned supervised learning. Our findings are in accordance with recent results utilizing generative language models for out-of-the-box extractive tasks.

\section{Conclusion}\label{sec:conclusion}

We introduced \system, a simple framework that constrains the inference of self-supervised language models after they have been trained. We perform an extensive experimental study over multiple relation extraction benchmarks and demonstrate that \system outperforms competing relation extraction methods by up to 27.8 $F_1$ points compared to the next-best unsupervised method.

\section*{Acknowledgments}

This work was supported by DARPA under grant
ASKE HR00111990013. The U.S. Government is authorized
to reproduce and distribute reprints for Governmental purposes notwithstanding any copyright notation thereon. Any
opinions, findings, and conclusions or recommendations expressed in this material are those of the authors and do not
necessarily reflect the views, policies, or endorsements, either expressed or implied, of DARPA or the U.S. Government.

\bibliography{re}
\bibliographystyle{acl_natbib}

\clearpage

\appendix
\begin{table*}[t]
\centering
\small
\begin{tabular}{l|l|l|l|l|l}
\textbf{Benchmark} & \textbf{Relation in Context} & \textbf{Extraction Type} & \textbf{Underlying Corpus} & \textbf{\# of Relations} & \textbf{Total Samples}\\\hline T-REx & \makecell[l]{Implicit Relation \\ Mention} & Single-Token & Wikipedia & 41 & 34,039\\\hline
Google-RE & \makecell[l]{Exact Relation \\ Mention} & Single-Token & Wikipedia & 3 & 5,528\\\hline
ZSRE & \makecell[l]{Possibly Irrelevant \\ Context} & \makecell[l]{Multi-Token} & Wikipedia & 120 & 42,635\\\hline
TACRED & \makecell[l]{Possibly Irrelevant \\ Context} & \makecell[l]{Multi-Token} & TAC-KBP & 41 & 6,357\\
\end{tabular}
\caption{We consider four benchmarks that vary with respect to the type of target extractions, the quality of context to relation alignment, and the underlying corpus.}
\label{tab:benchmarks}
\end{table*}

\section{Implementation Details}\label{sec:app:impl}
We set \system's top-$k$ parameter (see Section~\ref{sec:alignment}) to $16$. We tune the $\lambda$ parameter, when applicable, on the provided development sets of the datasets using the $F_1$ metric. Additionally, we tune whether to use the NER expansion, again using the development sets of the datasets. These hyperparameters are tuned using a standard grid search. We use Fairseq's implementation of RoBERTa-large\footnote{\url{https://github.com/pytorch/fairseq}} as our self-supervised language model. For the embeddings of the context rejection mechanism, we use the FastText library \cite{bojanowski2017enriching}. For the token embeddings of the anchor identification model, we first collect an embedding for each subword (RoBERTa uses byte-pair subword encodings \cite{sennrich2015neural}) by flattening the output representation of all of the RoBERTa-large decoder layers for each subword into a single vector. Because we operate on the token and not the subword level, we obtain a token representation by averaging all subword vector embeddings that compose a token. Examining the effect of our embedding choices is out of the scope of this work, and we leave it as a future analysis. 

As stated in our construction of $\tilde{B}$ (Section \ref{sec:alignment}), we filter any punctuation predicted. For named entity recognition and noun phrase chunking (used for identifying multi-token extractions in \system), we use the \texttt{en\_web\_core\_lg} model of the spaCy library\footnote{\url{https://spacy.io/}}. We train and run all models on a single NVIDIA V100 32GB memory GPU.

\begin{figure}[h]
\center
\includegraphics[width=0.5\textwidth]{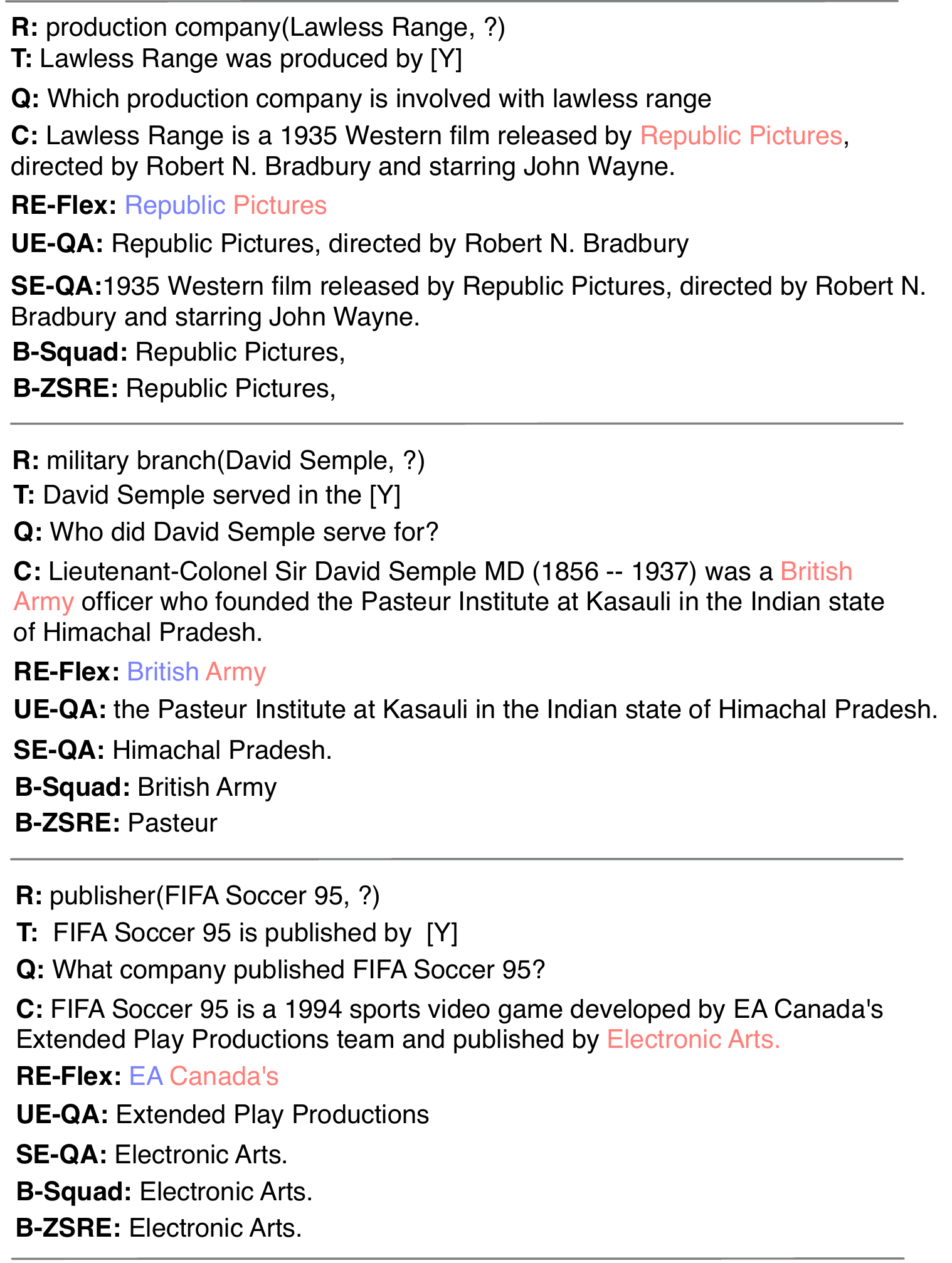}
\caption{Example extractions from ZSRE for the different methods. Here, \textbf{R} indicates the target relation, \textbf{T} the cloze template used, \textbf{Q} the corresponding question required by the QA-based models, and \textbf{C} the provided context for the extraction task.}
\centering
\label{fig:qual_examples}
\end{figure}

\section{Qualitative Results}\label{sec:qual}

We provide few qualitative examples of extractions from ZSRE obtained by the different methods. The examples are shown in Figure~\ref{fig:qual_examples}. The first two examples highlight two accurate extractions from \system, while the third example an incorrect extraction. These examples also highlight the weakness of the UE-QA and SE-QA methods: many times they extract an incorrect large sequence from the input context.

\section{Dataset Details}\label{sec:app:datasets}
\paragraph{TACRED adaptation to slot filling} Relation Extraction Dataset \cite{zhang2017position} is a relation classification dataset. The original task is to predict the relation of a subject and object pair given a supporting context. There are 41 possible relations, with an additional relation labelled ``no\_relation" to denote an example whose sentence does not express the relation between the subject and object. We convert the dataset to our slot filling setting by considering the subject and relation known for each example, and setting the task to predict the object. Following the established process of \citet{levy_zero-shot_2017} for adding realistic negative examples, we distribute all examples labelled \texttt{no\_relation} to relations sharing the same head entity, and set the target object for each to be $\varnothing$.

\paragraph{Dataset characteristics} A table of dataset characteristics can be found in Table \ref{tab:benchmarks}.

\section{Competing Methods Implementation Details}\label{sec:app:competingdetails}

All generative baselines are implemented using Fairseq \cite{ott2019fairseq}. Following the implementation of \cite{lewis_unsupervised_2019}, we train a BERT-Large model on the provided training datasets of \cite{lewis_unsupervised_2019} and \cite{dhingra_simple_2018} for the UE-QA and SE-QA baselines. These training datasets are collected over a snapshot of Wikipedia, which is the underlying corpus of three of our four benchmarks. We use the HuggingFace Transformers library \cite{wolf2019transformers} for our implementation of all QA models except BiDAF, for which we use a slightly altered version of the original author's code \cite{levy_zero-shot_2017}.

\section{Microbenchmarks}\label{sec:app:micro}
We evaluate the effect of different components of \system on its end-to-end performance. 

\newparagraph{Context rejection analysis} We first examine the effect of \system's context rejection mechanism. In Table~\ref{tab:context_ablation}, we measure the performance with and without context rejection on the datasets which require context rejection. We find that on the ZSRE dataset, the rejection increases $F_1$ by $3.5$. On TACRED, $F_1$ increases by $10.6$ $F_1$ with context rejection. In both cases, context rejection positively impacts performance.

\begin{table}
\centering
\small
\begin{tabular}{l|l l|l l}
\multirow{2}{*}{\textbf{Method}} & \multicolumn{2}{c|}{\textbf{ZSRE}} & \multicolumn{2}{c}{\textbf{TACRED}} \\ \cline{2-5}
& \textbf{EM} & \textbf{$\mathbf{F_1}$} & \textbf{EM} & \textbf{$\mathbf{F_1}$} \\ \hline
No expansion  & 42.4 & 46.1 & 49.6 & 50.3 \\ 
NER expansion  & 36.4 & 39.2 & 42.9 & 43.3 \\ 
Tuned expansion & 43.7 & 46.9 & 49.4 & 50.1 \\ 
\end{tabular}
\caption{Effect of token expansion on ZSRE dataset.}
\label{tab:tok_expansion}
\end{table}

\newparagraph{Anchor expansion analysis}
We examine the effect of expanding the anchor token in \system. To examine this behavior in more details, we evaluate \system by considering single-token only extractions, multi-token extractions using NER expansion, and a tuned expansion that chooses either to expand or not to expand based on performance on the development set for each dataset. The results are shown in Table~\ref{tab:tok_expansion}. We see that with tuned expansion, $F_1$ increases by $0.8$ $F_1$ on ZSRE, and decreases by $0.2$ $F_1$ on TACRED. In fact, utilizing NER expansion for all relations leads to a decrease of $6.9$ $F_1$ on ZSRE and $7.0$ $F_1$ on TACRED. We conclude that what additional information to include in a prediction is determined by the information need of each relation, and meeting this need for general relations is left for future work.

\begin{figure}
\center
\includegraphics[width=0.48\textwidth]{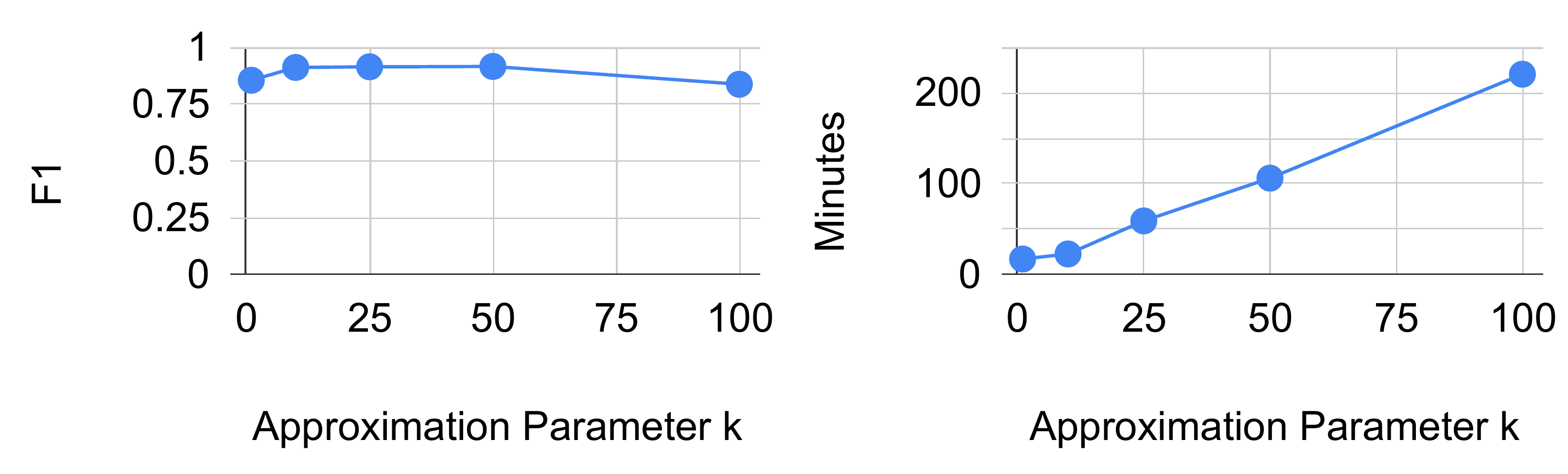}
\caption{Effect of parameter $k$ on $F_1$ for Google-RE.}
\centering
\label{fig:merged_topk}
\end{figure}

\newparagraph{Approximation analysis}
We examine the tradeoffs between performance, runtime, and the approximation parameter $k$ described in Section~\ref{sec:alignment}. We set the batch size to $1$ to for this analysis. Results for the three Google-RE relations are shown in Figure~\ref{fig:merged_topk}. Our measurements show that our choice of $k=16$ leads to high-quality results while having an acceptable runtime.

\section{Biases of QA Models}\label{sec:app:qabias}

Given that RE-Flex outperforms all supervised methods for T-Rex and Google-RE, we perform a detailed analysis to understand the reason behind this limitation of QA models. We suspect these results can be partially attributed to the construction of these settings, where the expected response is a single token; however QA models are more likely to predict multi-token spans because their training data is biased towards longer spans. 

We have the following finding from our results: B-ZSRE, which is trained on entity length answer spans, performs better than the B-Squad baseline by $17.6$ EM. As both models are the exact same architecture, but trained on different QA datasets, we can attribute this difference to biases in span length. We further verify this span length bias by conducting an error breakdown on these datasets. For each QA model, we consider each example which returns an EM of $0$, and classify the example based on whether the predict has no overlap with the ground truth, or by how much longer the prediction is.

\begin{figure}
\center
\includegraphics[width=0.48\textwidth]{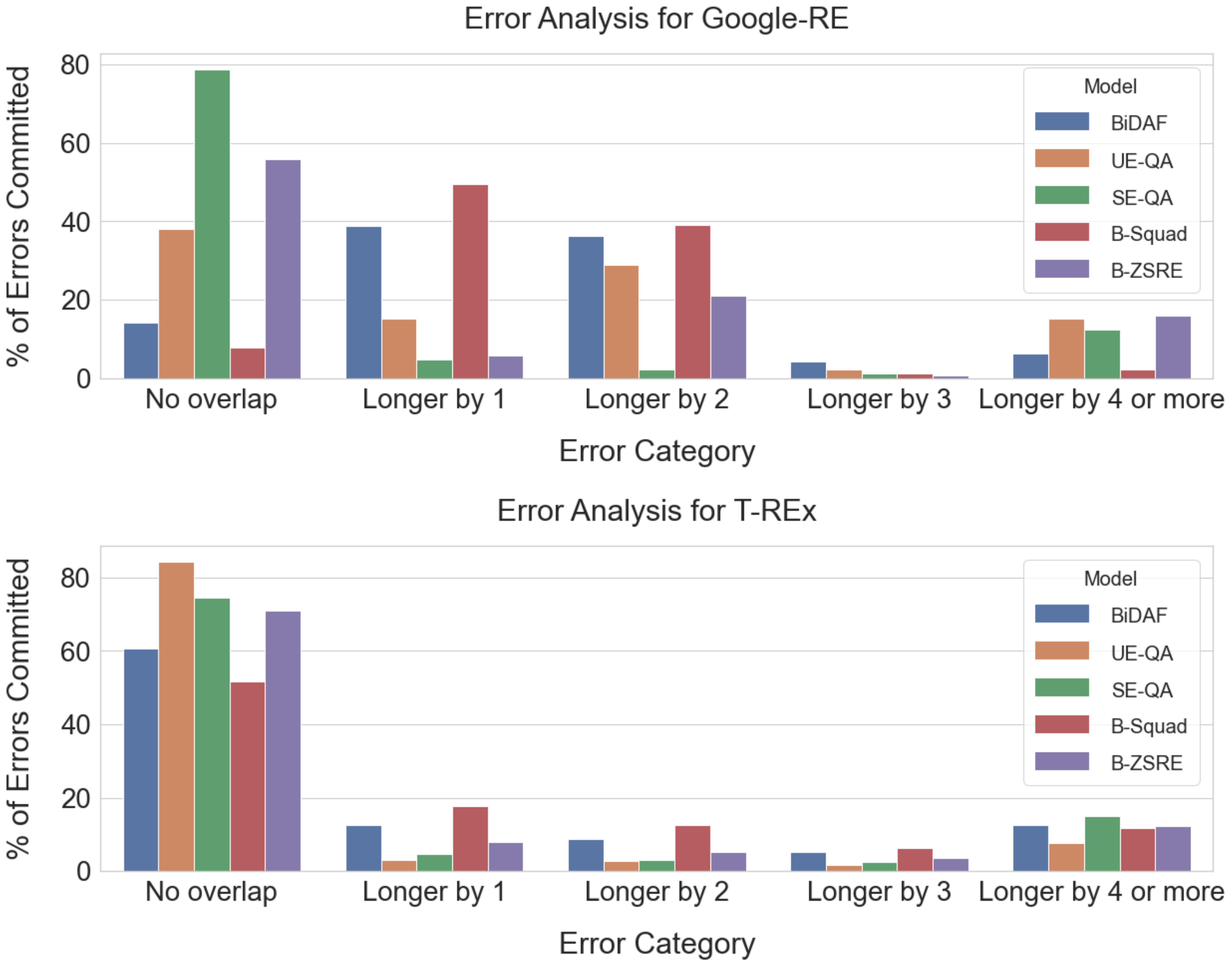}
\caption{Error categorization of QA-based methods. For \system all errors belong to the No overlap group.}
\centering
\label{fig:qa_googlre_trex_error}
\end{figure}

We present the results in Figure \ref{fig:qa_googlre_trex_error}. We see that on Google-RE, the majority of the errors committed by BiDAF and B-Squad, both trained on Squad 2.0, are because the predictions are longer than the expected answer by one or two tokens. B-ZSRE does not exhibit these error ratios, instead primarily missing the answer entirely. On T-REx, all models primarily miss the ground truth entirely. We attribute this finding to the fact that evidence in T-REx is weaker and does not have explicit lexical clues to select answer spans. Training these models using contexts with weaker evidence might improve relation extraction performance.

\noindent{\textbf{Takeaway:}} Supervised QA models are biased towards the span lengths in their training set, and struggle when given weaker evidence contexts.

\end{document}